\documentclass[10pt,twocolumn,letterpaper]{article}

\usepackage{iccv}
\usepackage{times}
\usepackage{epsfig}
\usepackage{graphicx}
\usepackage{amsmath}
\usepackage{amssymb}
\usepackage{epsfig}
\usepackage{graphicx}
\usepackage{amsmath}
\usepackage{amssymb}
\usepackage[ruled]{algorithm2e}
\usepackage{algorithmic}
\usepackage{multirow}
\usepackage{color}
\usepackage{color, colortbl}
\usepackage{graphicx}
\usepackage{algorithmic}
\usepackage{array}
\usepackage[english]{babel} % English language/hyphenation
\usepackage{amsmath,amsfonts,amsthm} % Math packages
\usepackage{pifont}
\usepackage{authblk}
\newcommand{\msign}{\operatorname{sign}}
\newcommand{\xmark}{\ding{55}}%

\usepackage[breaklinks=true,bookmarks=false]{hyperref}

\iccvfinalcopy % *** Uncomment this line for the final submission
\ificcvfinal\fi
\def\iccvPaperID{1648} % *** Enter the ICCV Paper ID here

\begin{document}

%%%%%%%%% TITLE
\title{Deep Binaries: Encoding Semantic-Rich Cues for Efficient Textual-Visual Cross Retrieval}
\author[1]{Yuming Shen
}
\author[1,2]{Li Liu
}
\author[1]{Ling Shao
}
\author[3]{Jingkuan Song
}
\vspace{-3ex}\affil[1]{School of Computing Sciences, University of East Anglia, Norwich, UK}
\affil[2]{ Malong Technologies Co., Ltd., Shenzhen, China}
\affil[3]{ Future Media Center, University of Electronic Science and Technology of China, Chengdu, China}
\affil[ ]{\small \texttt{yuming.shen@uea.ac.uk,  li.liu@malongtech.cn,  ling.shao@ieee.org,  jingkuan.song@gmail.com}}

\maketitle
\thispagestyle{empty}

%%%%%%%%% ABSTRACT
\begin{abstract}
Cross-modal hashing is usually regarded as an effective technique for large-scale textual-visual cross retrieval, where data from different modalities are mapped into a shared Hamming space for matching. Most of the traditional textual-visual binary encoding methods only consider holistic image representations and fail to model descriptive sentences. This renders existing methods inappropriate to handle the rich semantics of informative cross-modal data for quality textual-visual search tasks.

To address the problem of hashing cross-modal data with semantic-rich cues, in this paper, a novel integrated deep architecture is developed to effectively encode the detailed semantics of informative images and long descriptive sentences, named as Textual-Visual Deep Binaries (TVDB). In particular, region-based convolutional networks with long short-term memory units are introduced to fully explore image regional details while semantic cues of sentences are modeled by a text convolutional network. Additionally, we propose a stochastic batch-wise training routine, where high-quality binary codes and deep encoding functions are efficiently optimized in an alternating manner. Experiments are conducted on three multimedia datasets, i.e. Microsoft COCO, IAPR TC-12, and INRIA Web Queries, where the proposed TVDB model significantly outperforms state-of-the-art binary coding methods in the task of cross-modal retrieval.
\end{abstract}

%%%%%%%%% BODY TEXT
\section{Introduction}
\begin{figure}
	\begin{center}
		\includegraphics[width=0.95\linewidth, height=0.33\linewidth]{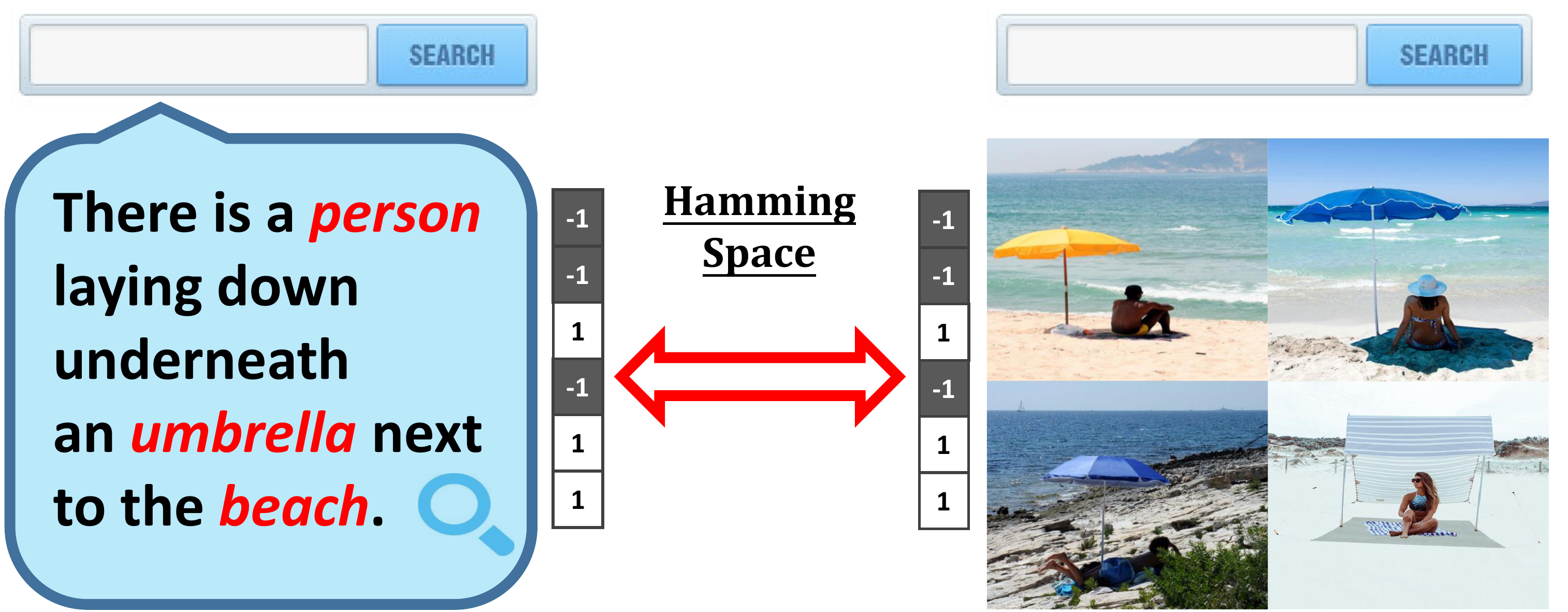}
		(a) TVDB (ours)
		\includegraphics[width=0.95\linewidth, height=0.33\linewidth]{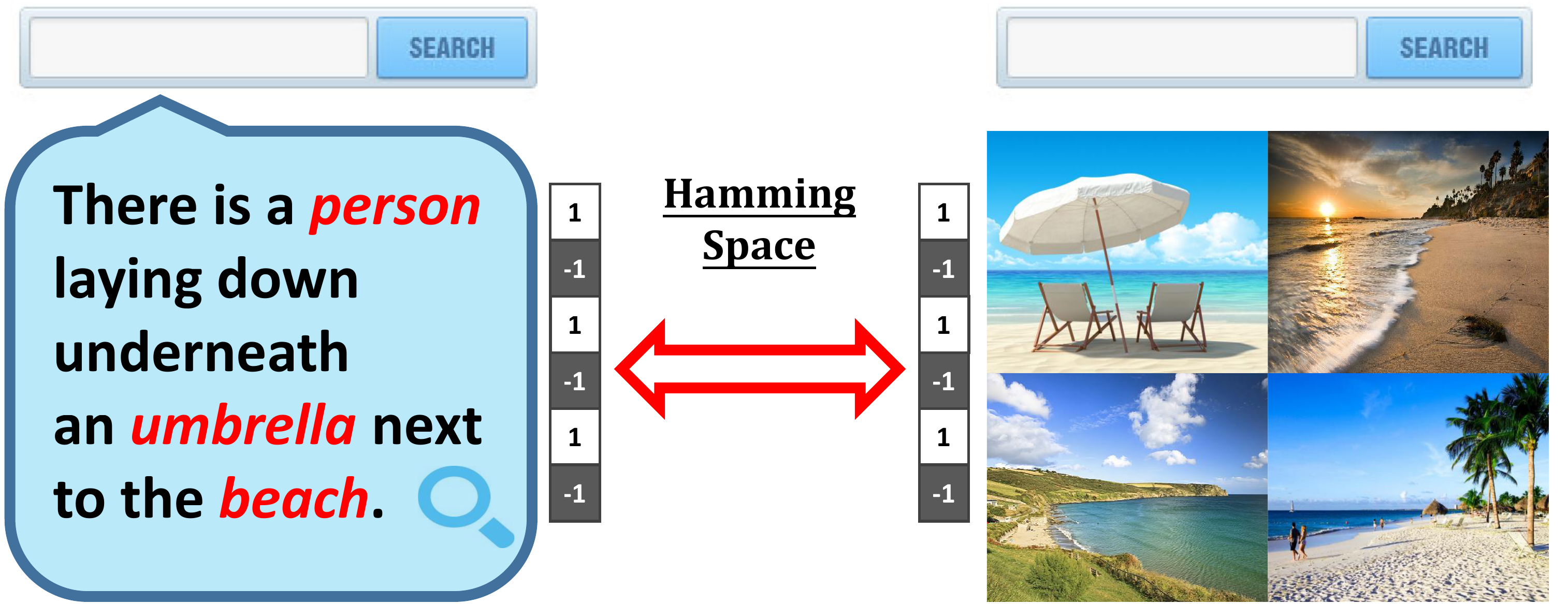}
		(b) Conventional Cross-Modal Hashing
		\vspace{1ex}
		\caption{The proposed model (a) aims at encoding all informative words of a sentence and all possible attractive regions in an image. Contrarily, conventional textual-visual binary encoding methods (b) only utilize simple representations of each modality and discard some information (\eg the instances of \textit{person} and \textit{umbrella} are not well-encoded), resulting in low matching quality.}
		\label{diff}
		\vspace{-2ex}
	\end{center}
\end{figure}
Data retrieval between image and text modalities has aroused a lot of recent attention, becoming an overwhelming research topic in computer vision. As deep learning technology develops, dramatic progress has been achieved recently in this area. %visual question answering \cite{malinowski2015ask,antol2015vqa,gao2015you}, caption generation \cite{xu2015show,donahue2015long,densecap} and
%textual-visual retrieval \cite{Xu:2015:SCD:2733373.2806346,Lu:2014:LMN:2647868.2655001,NIPS2014_5281,wang2015learning,frome2013devise}, with a variety of methods using real-valued shared space representations. %Promising performances have been achieved by densely encoding image details and language semantics. %Typically embedding different types of data with deep neural networks and trained with stochastic back propagation, these methods achieve impressive fine-grained multi-modal mapping performance \cite{Lu:2014:LMN:2647868.2655001,frome2013devise,NIPS2014_5281,wang2015learning, Jiang:2015:DCC:2733373.2806240}.
%However, due to the high complexity of similarity computation, real-valued embeddings are usually infeasible for large-scale textual-visual retrieval. Addressing this issue, 
Handling large-scale image-text retrieval, cross-modal hashing schemes \cite{cmfh,cvh,scm,imh,seph,qch,dvsh,dcmh,imvh,hth,plmh} have been proposed to encode heterogeneous data from a high-dimensional feature space into a shared low-dimensional Hamming space where an approximate nearest neighbour of a given query can be found efficiently.\vspace{1ex}

Traditionally, most existing researches in cross-modal hashing only focus on image-tag mapping \cite{cmfh,cvh,scm,imh,cmssh,seph,qch,dvsh,dcmh}, where holistic image representations and semantic tags feed the shallow binary coding procedure. It is insufficient to simply link images with tags, instead of real sentences, for multimedia searching problems; the detailed information of images is usually discarded due to the use of holistic representations and the semantics enclosed in descriptive sentences are hardly explored. 
The recent deep hashing methods \cite{dvsh,corrae,dcmh,cmnn} also suffer from several drawbacks. First of all, image data are likewise poorly modeled due to the lack of detailed regional information during encoding. We argue that this is not optimal for images with fruitful semantics. Secondly, most of the deep models mentioned above still utilize coarse text representations, which is inappropriate for modeling long sentences. Moreover, the network training efficiency can be further improved by more advanced code learning architectures.

Driven by the drawbacks of the previous works, in this paper, we consider a more challenging task to encode informative multi-modal data, \ie, semantic-rich images and descriptive sentences, into binary codes for cross-modal search, termed as Textual-Visual Deep Binaries (TVDB). Particularly, the popular Region Proposal Network (RPN) \cite{fasterrcnn} and Long Short-Term Memory (LSTM) \cite{lstm,lstm2} are introduced to formulate the image binary coding function, so that the regional semantic details in images can be well preserved from dominant to minor. Meanwhile, the latest advances in text Convolutional Neural Networks (CNN) \cite{tcnn,conneau2016very,NIPS2014_5550} are adopted to build the text binary encoding network, leveraging structural cues between the words in a sentence. The proposed deep architecture produces high-semantic-retentivity binary codes and achieves promising retrieval performance. 
%From this perspective, TVDB significantly differs from the existing deep textual-visual hashing methods. 
The intuitive difference between the proposed method and the traditional ones are given in Figure \ref{diff}. It can be seen that the proposed TVDB encodes as many details as possible from images and sentences, leading to more representative binary codes for matching.

%Driven by the drawbacks discussed above, we consider a more challenging task to encode informative multi-modal data, \ie scene-scale images and long descriptive sentences with rich information, into binaries for cross-modal search, termed as Deep Sentence-Vision Alignment (TVDB). Inspired by the recent success in deep hashing methods \cite{cao2016deep,xia2014supervised}, we apply deep learning architectures to the proposed model as hash functions so that both the detailed semantic retentivity and coding convergency can be guaranteed. It is worth noticing that Cao \etal \cite{dvsh} and Jiang \etal \cite{dcmh} also proposed cross-modal hashing with deep neural networks, namely DVSH and DCMH respectively. We deem that the proposed model is significantly diverse from DVSH and DCMH in terms of regional information processing capacity and training routine. In this paper, the popular Region-based Convolutional Neural Networks (R-CNN) \cite{fastrcnn,fasterrcnn} and Recurrent Neural Networks (RNN) \cite{lstm,lstm2} are introduced to formulate the image-side binary coding function, so that the regional semantic details can be well preserved. We also adopt the latest findings in text Convolutional Neural Networks (CNN) \cite{tcnn,conneau2016very,NIPS2014_5550} to build the text-side hashing network. The proposed network provide with high semantic encoding capacity and lead to promising retrieval results. The intuitive difference between our cross-modal mapping method and the traditional ones are given in Figure \ref{diff}.

\vspace{0.5ex}In addition to the novel deep binary encoding networks of TVDB, an efficient stochastic batch-wise code learning procedure is proposed. %, suitable for textual-visual retrieval tasks. 
Inspired by Shen \etal \cite{sdh}, the binary codes in TVDB are discretely and alternately optimized during the batch-wise learning procedure. 
Batching data randomly and iteratively, the proposed training routine guarantees an effective learning objective convergency.
% Our experiments also show this stochastic-batched training routine provides distinguishing improvement in performance compared with the existing training strategies.
The contributions of this work can be summarized as follows:\vspace{-0.5ex}
\begin{itemize}
	\item The TVDB model is proposed to effectively encode rich regional information of images as well as semantic dependencies and cues between words by exploiting two modal-specific deep binary embedding networks.  In this way, the intrinsic semantic correlation between heterogeneous data can be quantitatively measured and captured.\vspace{-0.5ex}
	%\item A novel stochastic batch-wise training strategy is designed to preserve binary constrains that provides high-quality supervisions to the deep neural networks and thus leads to quick converging.\vspace{-1ex}
	\item A novel stochastic batch-wise training strategy is adopted to optimize TVDB, in which reliable binary codes and the deep encoding functions are optimized in an alternating manner within every single batch.\vspace{-0.5ex}
	\item The evaluation results of our model on three semantically rich datasets highly surpass those of existing state-of-the-art binary coding methods in cross-modal retrieval.\vspace{-0.5ex}%, which reveals the supreme semantic encoding capacity of the proposed model.
\end{itemize}
\section{Related Work}
%Real-valued embeddings have been proved to be able to handle complex tasks between vision and language, \eg visual question answering \cite{malinowski2015ask,antol2015vqa,gao2015you}, caption generation \cite{xu2015show,donahue2015long,densecap}, and 

The cutting-edge studies in vision and language achieve promising results in terms of visual question answering \cite{malinowski2015ask,antol2015vqa,gao2015you}, caption generation \cite{xu2015show,donahue2015long,densecap}, and real-valued cross-modal retrieval \cite{JMLR:v15:srivastava14b,Lu:2014:LMN:2647868.2655001,NIPS2014_5281,wang2015learning,hu2015natural,scenegraph,fisher,dvsa,mao2014deep,ma2015multimodal,yan2015deep}. The best-performing real-valued cross-modal retrieval models typically rely on densely annotated image-region and text pairs for embedding. However, these methods are far from satisfactory for large-scale data retrieval due to the inefficient similarity computation of real-valued embeddings.

%On the other hand, existing research interests in textual-visual hashing %, generally termed as \textit{cross-modal retrieval}, usually focus on image-tag mapping. 
On the other hand, there exists several hashing methods \cite{liu2016sequential,weiss2009spectral,liu2011hashing,heo2012spherical,liu2016multilinear,gong2011iterative,raginsky2009locality,liu2012supervised,kulis2009kernelized,liu2017discretely} aiming at efficient retrieval. For textual-visual hashing, it has been a traditional and common solution to encode images and tags via shallow embedding functions with either unsupervised \cite{cvh,imh,cmfh,pdh,lcmh,msae,dmhor,lsph,liu2017sequential}, pairwise based \cite{cmssh,qch,imvh,rahh,hth,plmh,ccqq} or supervised \cite{smmh,cah,seph,scm,cmnn} code learning methods. 
%Among these traditional cross-modal binary encoding methods, SePH \cite{seph} has become a successful proposal in reducing data heterogeneity.% by transforming semantic affinities into a probability distribution model.
More recently, deep hashing methods \cite{cao2016deep,xia2014supervised,zhuang2016fast,deepsh,do2016learning,chn,liu2017deep} provide promising results in image recognition, which is also adopted in
%Cao \etal \cite{dvsh} and Jiang \etal \cite{dcmh}
\cite{dvsh,corrae,dcmh,cmnn,chn} for textual-visual retrieval. Jiang \etal proposed DCMH \cite{dcmh} for image-tag retrieval using a set of multi-layer neural networks which simply take deep holistic image features and word count vectors as input.
%In general, DCMH can be regarded as an extension of conventional cross-modal hashing methods with a slight improvement in terms of hash functions.
%is in the framework of supervised hashing, where label information is leveraged for training the cross-modal hash functions.
DVSH \cite{dvsh} proposed by Cao \etal becomes a more feasible solution for image-sentence hashing as the sequential information of sentence data is better encoded by introducing the Recurrent Neural Networks (RNN) \cite{lstm}. CDQ~\cite{chn} combines data representation learning steps with quantization error controlling hash coding methods with deep neural networks, while it still basically addresses image-tag hashing. In general, most of these methods can barely obtain adequate performance in our task due to the coarse image and text representations. %In the following sections, we will elaborate on the proposed TVDB model.

\vspace{-1.5ex}\section{Deep Encoding Networks for TVDB}\label{sec2}

\begin{figure*}[t]
	\begin{center}
		%\includegraphics[width=0.8\linewidth]{egfigure.eps}
		\includegraphics[width=0.99\linewidth, height=0.35\linewidth]{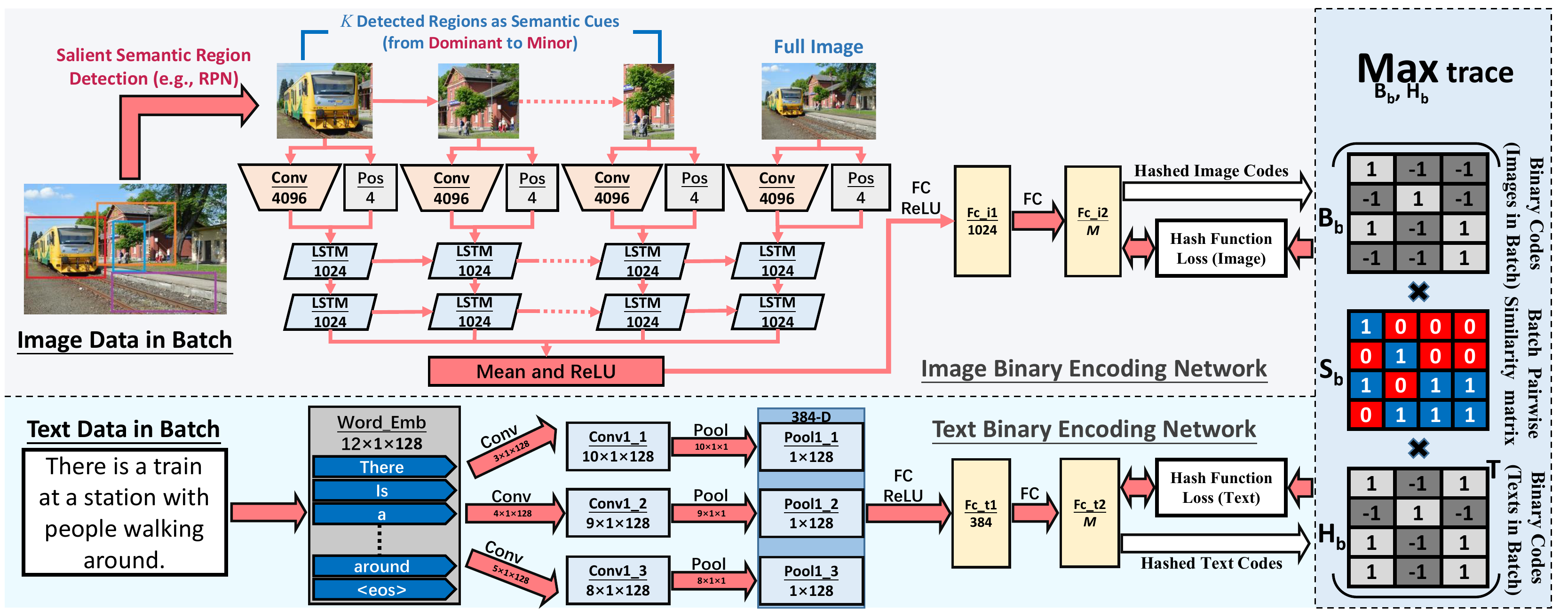}
	\end{center}
	\vspace{-1.5ex}
	\caption{The network architecture of TVDB. RPN, CNN and LSTM are utilized to form the image encoding function $\mathit{f}\left(\cdot\right)$ (the upper network), encoding image regions from dominant to minor. The sentence encoding network $\mathit{g}\left(\cdot\right)$ is built with a text-CNN (the lower network). The rightmost component refers to a \textbf{batch-wise} coding procedure. Here the module \texttt{Pos} computes the region coordinates discussed in subsection~\ref{s31}, while \texttt{Word\_Emb} refers to the linear word embedding procedure introduced in subsection \ref{s32}.}
	\label{archi}
	\vspace{-1.5ex}
\end{figure*}

%This work addresses the problem of data retrieval between informative images and long sentences using deep binary codes. As is shown in Figure \ref{archi}, the proposed TVDB model is basically composed of two deep neural networks and a batch-wise code learning phase, \ie, the rightmost component. The two deep neural networks play the role of binary encoding functions for images and sentences denoted as $\mathit{f}\left(\cdot\right)$ and $\mathit{g}\left(\cdot\right)$, of which the output is fixed to $M$ bits in length. 
%The batch-wise code learning phase optimizes the binary codes for all samples in the mini-batch as supervisions of $\mathit{f}\left(\cdot\right)$ and $\mathit{g}\left(\cdot\right)$ for efficient training.

This work addresses the problem of data retrieval between informative images and long sentences using deep binary codes. As shown in Figure \ref{archi}, the proposed TVDB model is composed of two deep neural networks and a batch-wise code learning phase. The two deep neural networks play the role of binary encoding functions for images and sentences denoted as $f\left(\cdot\right)$ and $\mathit{g}\left(\cdot\right)$ respectively. %, of which the output is fixed to $M$ bits in length. 
The batch-wise optimization allows using binary codes as supervision of $\mathit{f}\left(\cdot\right)$ and $\mathit{g}\left(\cdot\right)$ in a mini-batch during training.

Some preliminary notation is introduced here. We consider a multi-media data collection $\mathcal{O}=\{\mathbf{X}, \mathbf{Y}\}$ containing both image data $\mathbf{X}=\{\mathbf{x}_i\}_{i=1}^N$ and sentence data $\mathbf{Y}=\{\mathbf{y}_i\}_{i=1}^N$, with $N$ denoting the total number of data points of each modality in the dataset. The two deep binary encoding functions $\mathit{f}\left(\mathbf{X};\varTheta\right)$ and $\mathit{g}\left(\mathbf{Y};\varPhi\right)$ of TVDB are parameterized by $\mathbf\varTheta$ and $\varPhi$. The sign function is applied to $\mathit{f}\left(\cdot\right)$ and $\mathit{g}\left(\cdot\right)$ to produce binary representations:
\vspace{0.5ex}\begin{equation}\label{eq1}
\begin{split}
\mathbf{B}&=\operatorname{sign}\left(\mathit{f}\left(\mathbf{X};\varTheta\right)\right)\in\{-1,1\}^{M\times N}\text{,}\\
\mathbf{H}&=\operatorname{sign}\left(\mathit{g}\left(\mathbf{Y};\varPhi\right)\right)\in\{-1,1\}^{M\times N}\text{,}
\end{split}
\vspace{2ex}
\end{equation}
where $M$ refers to the target binary encoding length. In the following subsections, we introduce the setups of the deep binary encoding networks.%, following the two themes below.%\vspace{0.5ex}

\subsection{Image binary encoding network}\label{s31}

The architecture of the image encoding network $\mathit{f}\left(\cdot\right)$ is given in this subsection. %Nowadays an internet image usually carries a vast of information distributed to diverse regions of it.
As discussed previously, encoding the holistic image discards the informative patterns and produces poor coding quality, which is not desirable for our task. We consider a deep neural network architecture that embeds several salient regions of an image into a single binary vector to enrich the encoded semantic information. It is worthwhile to note that we are not directly linking every image region with a certain concept of a sentence as it is not feasible for cross-modal hashing problems and is contrary to the original intention of binary encoding in this work. Instead, regional semantic cues of images are leveraged here to improve the encoding quality.\vspace{1ex}% so that an image can be better mapped with sentences.

\textbf{Salient Semantic Region Selection.} As shown in Figure \ref{archi}, for each image in the mini-batch, TVDB firstly detects a number of regional proposals that possibly carry informative parts, \eg recognizable or dominative objects in the image.
Recent works in region-based CNN \cite{fastrcnn,fasterrcnn} show great potential in detecting semantically meaningful areas of an image. We adopt the framework of the state-of-the-art RPN \cite{fasterrcnn} as the proposal detection basis of TVDB. A total number of $K$ semantic regions are sampled for further processing according to a simple heuristic attraction score $a_k$ in descending order, that is:
\vspace{-1ex}\begin{equation}\vspace{-0.5ex}
a_k=(c_k+d_k)/2\text{,}
\end{equation}
where $c_k\in(0,1)$ denotes the confidence score determined by the RPN and $d_k\in(0,1)$ refers to the normalized proportion of the $k$-th detected proposal in an image. We consider those regional proposals with high attraction score $a_k$ semantically dominating to the whole image as they are usually the most recognizable image parts. This heuristic region selection solution highly fits the task of cross-modal binary encoding since it does not require any additional supervision or fine-grained region-sentence relations.% and provides good coding convergency.

\vspace{1ex}\textbf{Regional Representation and Augmentation.} The selected image regions are fed into CNNs to extract vectorized representations. The benchmark CNN architecture AlexNet \cite{cnn2} is involved here, from which a feature representation of 4096-D is obtained. To make the most of structural information, the feature vector of each region is augmented with four additional digits indicating the normalized height, width, and center coordinates of the corresponding region bounding box, making the whole regional representation a 5000-D vector for each.

\vspace{1ex}\textbf{Recurrent Network for Encoding.} For our task, it is desirable to use a method which capitalizes on information from the selected ordered regions so that dominating image parts contribute more to the final representation. 
It has been proved that human eyes sequentially browse image parts from dominant to minor \cite{rayner2009eye}.
Simulating this procedure, we sort the $K$ selected regional proposals according to their attraction scores in descending order and then the corresponding 5000-D representation for each proposal is sequentially fed into an RNN 
%so that these regional semantics can be combined for further encoding. 
so that the dominant image parts can be well utilized. Additionally, the CNN feature of the holistic image is also appended to the end of the RNN input sequence, making a total of $K+1$ semantic regions for encoding. In particular, a two-level Long Short-Term Memory (LSTM) \cite{lstm0} is implemented as the RNN unit with 1024-D output length, following the popular structure described in \cite{lstm}. Shown in Figure~\ref{archi}, the outputs of the LSTMs are averaged along the time sequence and appended with a ReLU \cite{relu} activation. Two fully-connected layers are applied to the top of the averaged LSTM outputs, with output dimension 1024 and $M$ respectively. Thus the whole image can be encoded into an $M$-bit binary vector using the $\operatorname{sign}\left(\cdot\right)$ function. We choose the identity function as the activation to the fully-connected layer for the convenience of code regression. %The network settings are given in Table \ref{tab1}. 
The convolutional layers for images are built following AlexNet~\cite{cnn2}.

\subsection{Sentence binary encoding network}\label{s32}

Although recurrent networks are widely adopted in textual-visual tasks \cite{malinowski2015ask,donahue2015long,densecap,grounding16}, it is still argued that RNNs such as LSTMs are not usually a superior choice for specific language tasks due to their non-structural designs \cite{NIPS2014_5550,conneau2016very}. We aim at encoding the structural and contextual cues between the words in a sentence to ensure the produced binary codes have adequate information capacity. To this end, the text-CNN~\cite{tcnn} is chosen as the text-side encoding network $\mathit{g}\left(\cdot\right)$, where each word in a descriptive sentence is firstly embedded into a word vector with a certain dimension and then convolution is performed along the word sequence. %This approach provides a hierarchical representation of the sentences.

We pre-process text data following the conventional manner where all sentences are appended with an \textit{eos} token and padded or truncated to a certain length with all full stops removed. The sentence length after preprocessing is fixed to $12$, which is about the mean length of text data in the datasets used in our experiments. Each word is embedded to a 128-D vector using a linear projection before being fed into the CNN. The text-CNN architecture in TVDB is similar to the one of Kim \cite{tcnn}, with more fully-connected layers for coding. The full configuration of our text-CNN is given at the bottom of Figure \ref{archi}. The \texttt{Word\_Emb} layer in Figure \ref{archi} refers to the word embedding, of which the parameters are also involved in the back-propagation (BP) procedure. For the text convolution setups, the first and second digits of the kernel size denote the height and width of the convolutional kernels, with the third digit being the kernel number. Note that, as text convolution is only performed along the word sequence, the second dimension of kernel size of the convolutional layers are always 1. In this work, the number of kernels for all convolutional layers is set to 128. We also build two fully-connected layers here, where the first one \texttt{Fc\_t1} takes inputs from all pooling layers followed by a ReLU activation, and for the second one, \texttt{Fc\_t2}, the identity activation is applied.

\section{Stochastic Batch-Wise Code Learning}
% The binary learning solution should be based on in-batch data similarities and provide reliable target binary codes as network supervision every time a mini-batch feeds. This ensures the whole framework works bet
The entire training procedure for $\mathit{f}\left(\cdot\right)$ and $\mathit{g}\left(\cdot\right)$ follows the mini-batch Stochastic Gradient Descent (SGD) since deep neural networks are utilized.
We suggest the binary learning solution should provide reliable target codes as network supervision every time a mini-batch feeds.  %To avoid biased training results, randomly shuffling needs to be performed frequently.

\subsection{Batch-wise alternating optimization}
Let $\mathcal{O}_{b}=\{\textbf{X}_{b},\textbf{Y}_{b}\}$ denote a mini-match stochastically taken from the data collection $\mathcal{O}$, where $\textbf{X}_{b}=\{\mathbf{x}_i\}_{i=1}^{N_b}$ and $\textbf{Y}_{b}=\{\mathbf{y}_i\}_{i=1}^{N_b}$ are image and sentence data in the mini-batch respectively. As the training process of TVDB is typically batch-based, we introduce an in-batch pair-wise similarity matrix $\mathbf{S}_b\in\{0,1\}^{N_b\times N_b}$ for target binary code learning, with $N_b$ denoting the number of data points within a mini-batch for training. The entry $\mathbf{s}_{pq}$ of $\mathbf{S}_b$ is defined as follows:
\vspace{-1ex}\begin{equation} \label{eqs}
\mathbf{s}_{pq}=\begin{cases}
1&{\mathbf{x}_p\text{, }\mathbf{y}_q\text{ share at least one sematic label,}}\\
0&\text{otherwise.}
\end{cases}
\end{equation}
The aim is to learn a set of target binary codes $\mathbf{B}_{b}$ for images and $\mathbf{H}_{b}$ for sentences that best describe the in-batch samples. Fully utilizing the pairwise relations $\mathbf{S}_b$ as supervision, a trace-based prototypic learning objective for hashing is thus built as
\vspace{0.5ex}\begin{equation}
\mathcal{L}\left(\mathbf{B}_b,\mathbf{H}_b,\mathbf{S}_b\right) =-\mathit{\operatorname{trace}}\left(\mathbf{B}_b\mathbf{S}_b\mathbf{H}_b^\top\right)\text{.}
\vspace{0.5ex}
\end{equation}
%This makes it possible for all related heterogeneous data to interact. 

%\textbf{Remark.} The trace-like objectives are also adopted in several recent works such as Deep CCA~\cite{yan2015deep}. We denote the main differences of TVDB from Deep CCA as follows: \textbf{1)} Deep CCA outputs real-valued representations for mapping while TVDB produces codes with binary constraints, which leads to different and more challenging ways to optimize the trace-like learning objectives; \textbf{2)} TVDB is a supervised hashing method, basing the similarity matrix $\mathbf{S}_b$ on label information, while Deep CCA computes the correlations according to the data representations.

A common learning procedure for cross-modal hashing can be formulated by solving the following problem:
\begin{equation}\label{obj1}\vspace{0.5ex}
\begin{split}
\min_{\mathbf{B}_b,\mathbf{H}_b,\varTheta,\varPhi}\mathcal{L}(\mathbf{B}_b,&\mathbf{H}_b,\mathbf{S}_b)\text{,}\\
\text{s.t. }\mathbf{B}_b=\msign\left(\mathit{f}\left(\mathbf{X}_b;\varTheta\right)\right),\,& \mathbf{H}_b=\msign\left(\mathit{g}\left(\mathbf{Y}_b;\varPhi\right)\right)\text{.}
\end{split}
\vspace{0.5ex}
\end{equation}
Relaxing the binary constraints to be continuous, \ie, $\mathbf{B}_b=\mathit{f}\left(\mathbf{X}_b;\varTheta\right), \mathbf{H}_b=\mathit{g}\left(\mathbf{X}_b;\varPhi\right)$, results in a slow and difficult optimization process. Inspired by Shen \etal \cite{sdh}, we reformulate the problem of (\ref{obj1}) by keeping the binary constraints and regarding $\mathbf{B}_b$ and $ \mathbf{H}_b$ as auxiliary variables,
\vspace{1ex}\begin{equation}\label{obj2}
\begin{split}
\min_{\mathbf{B}_b,\mathbf{H}_b,\varTheta,\varPhi} &\mathcal{L}\left(\mathbf{B}_b,\mathbf{H}_b,\mathbf{S}_b\right)+\eta\big(\lVert\mathbf{B}_b-\mathit{f}\left(\mathbf{X}_b;\varTheta\right)\rVert_{\text{F}}^2\\
+&\lVert\mathbf{H}_b-\mathit{g}\left(\mathbf{Y}_b;\varPhi\right)\rVert_{\text{F}}^2\big)\text{,}\\
\text{s.t. }\mathbf{B}_b&\in\{-1,1\}^{M\times N_b}, \mathbf{H}_b\in\{-1,1\}^{M\times N_b}\text{,}
\end{split}
\vspace{1ex}
\end{equation}
where $\eta$ is a penalty hyper parameter and $\lVert\cdot\rVert_{\text{F}}$ refers to the Frobenius norm. The two Frobenius norms here depict the quantization error between the binary codes $\mathbf{B}_b,\text{ }\mathbf{H}_b$ and the binary coding function outputs $\mathit{f}\left(\cdot\right),\text{ }\mathit{g}\left(\cdot\right)$. It has been proved in \cite{sdh} that with a sufficiently large value of $\eta$, Eq. (\ref{obj2}) becomes a close approximation to  Eq.~(\ref{obj1}), in which slight disparities between $\mathbf{B}_b$ and $\mathit{f}\left(\mathbf{X}_b;\varTheta\right)$ or $\mathbf{H}_b$ and $\mathit{g}\left(\mathbf{Y}_b;\varTheta\right)$ are tolerant to our binary learning problem.

Therefore, the comprehensive learning objective of TVDB is formulated. We provide optimization schemes below. It is observed that Eq.~(\ref{obj2}) is a non-convex NP-hard problem due to the binary constraints.
To better access it, an alternating solution based on coordinate descent is adopted to sequentially optimize $\mathbf{B}_b$, $\mathbf{H}_b$, $\varTheta$ and $\varPhi$ in every single batch as follows.

%By alternatively updating $\mathbf{B}_b$, $\mathbf{B}_b$, $\varTheta$ and $\varPhi$, problem (\ref{obj2}) can be solved with a reasonable learning objective $\mathcal{L}$.\vspace{0.5ex} Then we define a set of alternating steps to solve problem (\ref{obj2}).\vspace{1.5ex}

\textbf{Updating $\mathbf{B}_b$.} By fixing $\mathbf{H}_b$, $\varTheta$ and $\varPhi$, the subproblem of (\ref{obj2}) \wrt $\mathbf{B}_b$ can be written as\vspace{1ex}
\begin{equation}
\begin{split}
\min_{\mathbf{B}_b}\text{ }&\eta\lVert\mathbf{B}_b-\mathit{f}\left(\mathbf{X}_b;\varTheta\right)\rVert_{\text{F}}^2 -\mathit{\operatorname{trace}} \left(\mathbf{B}_b\mathbf{S}_b\mathbf{H}_b^\top\right)\text{,}\\
\text{s.t. }&\mathbf{B}_b\in\{-1,1\}^{M\times N_b}\text{,}\\
\end{split}
\end{equation}
to which the closed-form optimal solution becomes\vspace{1ex}
\begin{equation}\label{updatedb}
\mathbf{B}_b=\mathit{\operatorname{sign}}\left(2\eta f\left(\mathbf{X}_b;\varTheta\right)+\mathbf{S}_b\mathbf{H}_b^\top\right)\text{.}
\end{equation}

\vspace{1.5ex}\textbf{Updating $\mathbf{H}_b$.} Similar to $\mathbf{B}_b$, the solution to the subproblem of (\ref{obj2}) \wrt $\mathbf{H}_b$ is
\begin{equation}\label{updatedh}
\mathbf{H}_b=\mathit{\operatorname{sign}}\left(2\eta g\left(\mathbf{Y}_b;\varTheta\right)+\mathbf{B}_b{\mathbf{S}_b}\right)\text{.}
\end{equation}

\vspace{1.5ex}\textbf{Updating $\varTheta$ and $\varPhi$.} The subproblem of (\ref{obj2}) \wrt $\varTheta$ and \wrt $\varPhi$ can be respectively written as\vspace{1.5ex}
\begin{equation}\label{loss}
\begin{split}
\min_{\varTheta}&{\text{ }\operatorname{loss}_f:=\eta\lVert\mathbf{B}_b-\mathit{f}\left(\mathbf{X}_b;\varTheta\right)\rVert_{\text{F}}^2}\text{,}\\
\min_{\varPhi}&{\text{ }\operatorname{loss}_g:=\eta\lVert\mathbf{H}_b-\mathit{g}\left(\mathbf{Y}_b;\varPhi\right)\rVert_{\text{F}}^2}\text{,}
\end{split}
\end{equation}
when all other variables are fixed. These two subproblems are typically in the form of the $l2$ norm which are differentiable problems and thus  $\varTheta$ or $\varPhi$ can be optimized in the framework of SGD using back-propagation. Here the updated auxiliary binary codes $\mathbf{B}_b$ and $\mathbf{H}_b$ act as the supervisions to the binary encoding networks $\mathit{f}\left(\cdot\right)$ and $\mathit{g}\left(\cdot\right)$.% respectively.\vspace{0.5ex}

\subsection{Stochastic-batched training procedure.} 
In the above subsection, we have presented the binary code learning algorithm for each mini-batch of TVDB. However, how to apply the batch-wise learning objective to the whole dataset has not yet been discussed. In this subsection, the overall training procedure for mini-batch SGD is introduced.
%Eq. (\ref{updatedb}) and (\ref{updatedh}) are able to produce high-quality target binary codes for the encoding functions according to instant $\varTheta$, $\varPhi$ and $\mathbf{S}_b$, whilst the loss function of our binary encoding networks are also variant to the values of $\mathbf{S}_b$. 
Note that simply keeping data in every mini-batch unaltered in each training epoch usually results in poorly-learned hash functions. This is because the cross-modal in-batch data are not able to interact with the data outside of the batch and thus the batch-wise similarity $\mathbf{S}_b$ skews the statistics of the whole dataset.
To be more precise, we consider a data batching scheme to build every $\mathcal{O}_b$ and $\mathbf{S}_b$ that well explores the cross-modal semantic relationships across the entire dataset. To this end, a stochastic batching routine is designed. Each data mini-batch is randomly formed before being input into the training procedure. Therefore, $\mathbf{S}_b$ varies across each batch, which ensures the in-batch data diversity. %We figure out that this data batching scheme is suitable for our learning objective.

\begin{algorithm}[t]
	\caption{The Training Process of TVDB}
	\label{alg}
	%	\begin{algorithmic}
	\textbf{Input:}\hspace{0mm} Image-sentence dataset $\mathcal{O}=\{\mathbf{X},\mathbf{Y}\}$, Max \\
	$\quad\text{  }\quad$ training iteration $T$\\
	\textbf{Output:}\hspace{0mm} Hash function parameters $\varTheta$ and $\varPhi$\\
	\BlankLine
	Randomly initialize $\mathbf{B},\,\mathbf{H}\in\{-1,1\}^{M\times N}$ \\
	%Randomly initialize $\mathbf{H}\in\{-1,1\}^{M\times N}$\\
	\Repeat{convergence or max training iter $T$ is reached}{
		
		Get a \textbf{stochastic} mini-batch $\mathcal{O}_{b}$ from $\mathcal{O}$\\
		Get $\mathbf{B}_b,\,\mathbf{H}_b$ from $\mathbf{B},\,\mathbf{H}$  with respective indices\\
		Build $\mathbf{S}_b$ according to data relations and labels\\
		\textbf{Update $\mathbf{B}_b$} $\leftarrow$ Eq.(\ref{updatedb})\\
		\textbf{Update $\mathbf{H}_b$} $\leftarrow$ Eq.(\ref{updatedh})\\
		$(\operatorname{loss}_f,\operatorname{loss}_g)\leftarrow$ Eq.(\ref{loss})\\
		\textbf{Update}\\ $\quad(\varTheta,\varPhi)\leftarrow\left(\varTheta,\varPhi\right)-\mathbf{\Gamma}\left(\nabla_\varTheta\operatorname{loss}_f,\nabla_\varPhi\operatorname{loss}_g\right)$
		%		\textbf{Reshuffle} $\mathbf{D}, \mathbf{B}, \mathbf{H}$
	}
	%	\end{algorithmic}}
	\vspace{-1mm}
\end{algorithm}

Combining the stochastic batching method with the alternating parameter updating schemes, the whole training procedure of TVDB is illustrated in Algorithm \ref{alg}. The operator $\mathbf{\Gamma}\left(\cdot\right)$ in Algorithm \ref{alg} indicates the adaptive gradient scaler used for SGD, which is the Adam optimizer \cite{adam} in this work. Unlike some existing deep hashing methods \cite{dcmh} which update the target binary codes when a whole epoch is finished% so that the codes can be only used for the next training epoch
, TVDB updates $\mathbf{B}_b$ and $\mathbf{H}_b$ instantly when a mini-batch arrives. This training routine proves to achieve fast convergency for effectively learning the encoding networks $\mathit{f}\left(\cdot\right)$ and $\mathit{g}\left(\cdot\right)$ as shown in Figure~\ref{figeff}. 
%The code learning procedure of TVDB differs from those of recent deep hashing methods. For instance, DCMH~\cite{dcmh} updates codes only after each epoch of training. %In general, the alternating parameter updating manner permits the qualities of the target binary codes and the stochastic batching routine guarantees the overall training consistency. This makes the proposed stochastic batch-wise code learning routine successful in dealing with cross-modal deep binary encoding. 
Once the TVDB model is trained, given an image query $\mathbf{x}_q$ for example, we compute its binary code by $\mathbf{b}_q=\operatorname{sign}\left({f}\left(\mathbf{x}_q;\varTheta\right)\right)$. While for the retrieval database, 
the unified binary codes from each sentence is obtained via $\mathbf{H}=\operatorname{sign}\left(\mathit{g}\left(\mathbf{Y};\varPhi\right)\right)$. A sentence query can be processed in the similar manner.
%-------------------------------------------------------------------------
\section{Experiments}
\begin{table*}
	\vspace{-0.5ex}
	\begin{center}
		\caption{Image-sentence cross-modal retrieval MAP results of the proposed model compared with existing methods on the three datasets.}
		\label{cmp1}
		\small
		\resizebox{0.98\textwidth}{!}{
			\begin{tabular}{|c|c|c||cccc||cccc||cccc|}
				\hline
				\multirow{2}{*}{\textbf{Task}}&\multirow{2}{*}{\textbf{Method}}&\multirow{1}{*}{\textbf{Binary}}&\multicolumn{4}{c||}{\textbf{Microsoft COCO}}&\multicolumn{4}{c||}{\textbf{IAPR TC-12}}&\multicolumn{4}{c|}{\textbf{INRIA Web Queries}}\\\cline{4-15}
				&&\textbf{Code}& 16 bits & 32 bits & 64 bits & 128 bits & 16 bits & 32 bits & 64 bits & 128 bits & 16 bits & 32 bits & 64 bits & 128 bits  \\ \hline\hline
				
				\multirow{16}{*}{\parbox[t]{0.12\textwidth}{\centering Image Query\\ Sentence }}& CVFL~\cite{cvfl}&\xmark&\multicolumn{4}{c||}{0.488 (4096-D)}&\multicolumn{4}{c||}{0.499 (4096-D)}&\multicolumn{4}{c|}{0.137 (4096-D)}\\
				& PLSR~\cite{plsr}&\xmark&\multicolumn{4}{c||}{0.528 (4096-D)}&\multicolumn{4}{c||}{0.485 (4096-D)}&\multicolumn{4}{c|}{0.258 (4096-D)}\\
				& CCA~\cite{cca}&\xmark& 0.544& 0.536& 0.508& 0.469& 0.468& 0.434& 0.406& 0.389& 0.204& 0.217& 0.189& 0.152\\\cline{2-15}
				& CMFH~\cite{cmfh}&\checkmark& 0.488& 0.506& 0.508& 0.383& 0.442& 0.437& 0.361& 0.362& 0.197& 0.236& 0.245& 0.238\\
				& CVH~\cite{cvh}&\checkmark& 0.489& 0.474& 0.436& 0.402& 0.422& 0.401& 0.384& 0.374& 0.204& 0.200& 0.187& 0.163\\
				& SCM~\cite{scm}&\checkmark& 0.529& 0.563& 0.587& 0.601& 0.486& 0.505& 0.515& 0.522& 0.115& 0.215& 0.271& 0.308\\
				& IMH~\cite{imh}&\checkmark& 0.615& 0.650& 0.657& 0.677& 0.463& 0.490& 0.510& 0.521& 0.233& 0.250& 0.277& 0.295\\
				& QCH~\cite{qch}&\checkmark& 0.572& 0.595& 0.613& 0.634& 0.526& 0.555& 0.579& 0.605&-&-&-&-\\
				& CMSSH~\cite{cmssh}&\checkmark& 0.405& 0.489& 0.441& 0.448& 0.345& 0.337& 0.348& 0.374& 0.151& 0.162& 0.155& 0.151\\
				& SePH~\cite{seph}&\checkmark& 0.581& 0.613& 0.625& 0.634& 0.507& 0.513& 0.515& 0.530& 0.126& 0.137& 0.134& 0.142\\\cline{2-15}
				& CorrAE~\cite{corrae}&\checkmark& 0.550& 0.556& 0.570& 0.580& 0.495& 0.525& 0.558& 0.589&-&-&-&-\\
				& CM-NN~\cite{cmnn}&\checkmark& 0.556& 0.560& 0.585& 0.594& 0.516& 0.542& 0.577& 0.600&-&-&-&-\\
				& DNH-C~\cite{dnhc}&\checkmark& 0.535& 0.556& 0.569& 0.582& 0.480& 0.509& 0.526& 0.535&-&-&-&-\\
				& DCMH~\cite{dcmh}&\checkmark& 0.562& 0.597& 0.609& 0.646& 0.443& 0.491& 0.559& 0.556& 0.241& 0.268& 0.301& 0.384\\
				& DVSH~\cite{dvsh}&\checkmark& 0.587& 0.713& 0.739& 0.755& 0.570& 0.632& 0.696& 0.724&-&-&-&-\\\cline{2-15}
				&\textbf{TVDB}&\checkmark&\textbf{0.702}&\textbf{0.781}&\textbf{0.797}&\textbf{0.818}&\textbf{0.629}&\textbf{0.697}&\textbf{0.731}&\textbf{0.772}&\textbf{0.368}&\textbf{0.405}&\textbf{0.419}&\textbf{0.446}\\\hline\hline
				\multirow{16}{*}{\parbox[t]{0.12\textwidth}{\centering Sentence\\Query Image }}& CVFL~\cite{cvfl}&\xmark&\multicolumn{4}{c||}{0.561 (4096-D)}&\multicolumn{4}{c||}{0.495 (4096-D)}&\multicolumn{4}{c|}{0.291 (4096-D)}\\
				& PLSR~\cite{plsr}&\xmark&\multicolumn{4}{c||}{0.538 (4096-D)}&\multicolumn{4}{c||}{0.492 (4096-D)}&\multicolumn{4}{c|}{0.257 (4096-D)}\\
				& CCA~\cite{cca}&\xmark& 0.545& 0.542& 0.513& 0.468& 0.471& 0.438& 0.413& 0.395& 0.212& 0.228& 0.188& 0.150\\\cline{2-15}
				& CMFH~\cite{cmfh}&\checkmark& 0.574& 0.507& 0.510& 0.472& 0.447& 0.445& 0.438& 0.365& 0.178& 0.231& 0.248& 0.131\\
				& CVH~\cite{cvh}&\checkmark& 0.486& 0.470& 0.434& 0.401& 0.424& 0.403& 0.386& 0.376& 0.204& 0.200& 0.179& 0.164\\
				& SCM~\cite{scm}&\checkmark& 0.523& 0.544& 0.569& 0.585& 0.495& 0.514& 0.523& 0.529& 0.123& 0.226& 0.297& 0.349\\
				& IMH~\cite{imh}&\checkmark& 0.610& 0.679& 0.728& 0.740& 0.516& 0.526& 0.534& 0.527& 0.251& 0.275& 0.306& 0.284\\
				& QCH~\cite{qch}&\checkmark& 0.574& 0.606& 0.638& 0.667& 0.500& 0.536& 0.565& 0.589&-&-&-&-\\
				& CMSSH~\cite{cmssh}&\checkmark& 0.375& 0.384& 0.340& 0.360& 0.363& 0.377& 0.365& 0.348& 0.154& 0.153& 0.156& 0.147\\
				& SePH~\cite{seph}&\checkmark& 0.613& 0.649& 0.672& 0.693& 0.471& 0.480& 0.481& 0.495& 0.226& 0.256& 0.291& 0.319\\\cline{2-15}
				& CorrAE~\cite{corrae}&\checkmark& 0.559& 0.581& 0.611& 0.626& 0.498& 0.520& 0.533& 0.550&-&-&-&-\\
				& CM-NN~\cite{cmnn}&\checkmark& 0.579& 0.598& 0.620& 0.645& 0.512& 0.539& 0.549& 0.565&-&-&-&-\\
				& DNH-C~\cite{dnhc}&\checkmark& 0.525& 0.559& 0.590& 0.634& 0.469& 0.484& 0.491& 0.505&-&-&-&-\\
				& DCMH~\cite{dcmh}&\checkmark& 0.595& 0.601& 0.633& 0.658& 0.486& 0.487& 0.499& 0.541& 0.227& 0.305& 0.322& 0.380\\
				& DVSH~\cite{dvsh}&\checkmark& 0.591& 0.737& 0.758& 0.767& 0.604& 0.640& 0.680& 0.675&-&-&-&-\\\cline{2-15}
				&\textbf{TVDB}&\checkmark&\textbf{0.713}&\textbf{0.779}&\textbf{0.787}&\textbf{0.810}&\textbf{0.674}&\textbf{0.678}&\textbf{0.704}&\textbf{0.721}&\textbf{0.353}&\textbf{0.462}&\textbf{0.464}&\textbf{0.470}\\\hline
			\end{tabular}
		}
		\vspace{1ex}%\scriptsize{Note that CCA, CVFL and PLSR are real-valued methods. CorrAE, CM-NN, DNH-C, DCMH and DVSH are deep hashing methods.}
	\end{center}
	\vspace{-5ex}
\end{table*}
Experiments of TVDB on cross-modal retrieval are performed on three semantically fruitful sentence-vision datasets: Microsoft COCO \cite{coco}, IAPR TC-12 \cite{iapr} and INRIA Web Queries \cite{inria}. The evaluation results are reported according to the following themes: \textbf{(a) comparison with state-of-the-arts methods}, \textbf{(b) deep encoding network ablation study} and \textbf{(c) training routine feasibility}.
%\begin{itemize}
%	\item{{Comparison with Other Methods.}}\vspace{-0.5ex} %We compare the proposed TVDB algorithm with some state-of-the-art image-text hashing methods to demonstrate the overall high-quality retrieval performance of it.}\vspace{-0.5ex}
%	\item{{Deep Encoding Network Plausibility.}}\vspace{-0.5ex} %To prove that the deep encoding networks in this paper are designed reasonably, a set of variants of TVDB are built as baselines by modifying the deep encoding networks with some other architectures. }\vspace{-0.5ex}%Comparisons show the proposed deep networks of TVDB are successfully designed.}
%	\item{{Training Routine Feasibility.}}\vspace{-0.5ex} %The training efficiency and convergency of the proposed \textit{stochastic batch-wise binary code learning routine} are illustrated. }%Again, a series of baselines of TVDB with some shallowed training routines are compared.}
%\end{itemize}

For implementation details, we utilize the RPN \cite{fasterrcnn} for image proposal detection to pick $K=20$ informative regions for further LSTM-based encoding, and the value of $\eta$ in problem (\ref{obj2}) is set to $10^{-4}$ via cross-validation. %The impact of hyper parameters $\eta$ and $K$ are analysed in the \textbf{supplementary document}, where the time complexity of TVDB is also illustrated.
%The value of $K$ is set to $K=20$ for best performance.
%The parameters of Faster R-CNN model are directly obtain from the pre-trained model in \cite{fasterrcnn} in this paper and not involved in the procedure of BP for the convenience of training.
For the image-side CNNs, AlexNet \cite{cnn2} without its \texttt{fc\_8} layer is adopted with the pre-trained parameters from ImageNet classification \cite{imagenet}. The TVDB framework is implemented using Tensorflow~\cite{tf}. %\footnote{https://www.tensorflow.org/}.% and trained on an NVIDIA TITAN X GPU.

\subsection{Experimental settings}
The experiments of sentence-vision retrieval are taken on three multimedia datasets. Following the conventional textual-visual retrieval measures \cite{seph, dvsh}, the relevant instances for a query are defined by sharing at least one label.

\vspace{1ex}\textbf{Microsoft COCO \cite{coco}.} The COCO dataset contains a training image set of $80,000$ samples with about $40,000$ validation images. Each image is assigned five sentence descriptions and labeled with 80 semantic topics. To be consistent with \cite{dvsh}, we randomly select 5,000 images from the validation set and thus the retrieval gallery becomes around 85,000 images, from which we explicitly take 5,000 pairs as the query set and 50,000 images for training.
\begin{table*}[t]
	\begin{center}
		\caption{Ablation study: cross-modal retrieval MAP comparison of TVDB with several deep network baselines.}
		\label{cmp2}
		\small
		\resizebox{0.94\textwidth}{!}{
			\begin{tabular}{|c|c||c|c|c|c||c|c|c|c|}
				\hline
				\multirow{2}{*}{\textbf{Task}}&\multirow{2}{*}{\textbf{Method}}&\multicolumn{4}{c||}{\textbf{Microsoft COCO}}&\multicolumn{4}{c|}{\textbf{IAPR TC-12}}\\\cline{3-10}
				&& 16 bits & 32 bits & 64 bits & 128 bits & 16 bits & 32 bits & 64 bits & 128 bits\\ \hline\hline
				
				\multirow{5}{*}{\parbox[t]{0.12\textwidth}{\centering Image Query\\ Sentence }}&
				TVDB-I1 (full image only)& 0.565& 0.597& 0.621& 0.679& 0.468& 0.526& 0.597& 0.589\\
				& TVDB-I2 (ave-pooled regions)& 0.677& 0.692& 0.704& 0.726& 0.533& 0.572& 0.647& 0.656\\
				& TVDB-T1 (text bag-of-words)& 0.553& 0.565& 0.591& 0.631& 0.436& 0.508& 0.503& 0.548\\
				& TVDB-T2 (text LSTM)& 0.654& 0.717& 0.742& 0.775& 0.551& 0.619& 0,679& 0.726\\\cline{2-10}
				&\textbf{TVDB (full model)}&\textbf{0.702}&\textbf{0.781}&\textbf{0.797}&\textbf{0.818}&\textbf{0.629}&\textbf{0.678}&\textbf{0.731}&\textbf{0.772}\\\hline\hline
				\multirow{5}{*}{\parbox[t]{0.12\textwidth}{\centering Sentence\\Query Image }}&
				TVDB-I1 (full image only)& 0.573& 0.634& 0.653& 0.657& 0.462& 0.514& 0.516& 0.568\\
				& TVDB-I2 (ave-pooled regions)& 0.633& 0.645& 0.687& 0.717& 0.541& 0.568& 0.615& 0.629\\
				& TVDB-T1 (text bag-of-words)& 0.554& 0.609& 0.613& 0.654& 0.508& 0.516& 0.553& 0.590\\
				& TVDB-T2 (text LSTM)& 0.642& 0.734& 0.761& 0.782& 0.613& 0.606& 0.635& 0.696\\\cline{2-10}
				&\textbf{TVDB (full model)}&\textbf{0.713}&\textbf{0.779}&\textbf{0.787}&\textbf{0.810}&\textbf{0.674}&\textbf{0.697}&\textbf{0.704}&\textbf{0.721}\\\hline
			\end{tabular}
		}
	\end{center}
	\vspace{-2ex}
\end{table*}
\begin{figure*}
	\begin{center}
		\includegraphics[width=0.975\textwidth]{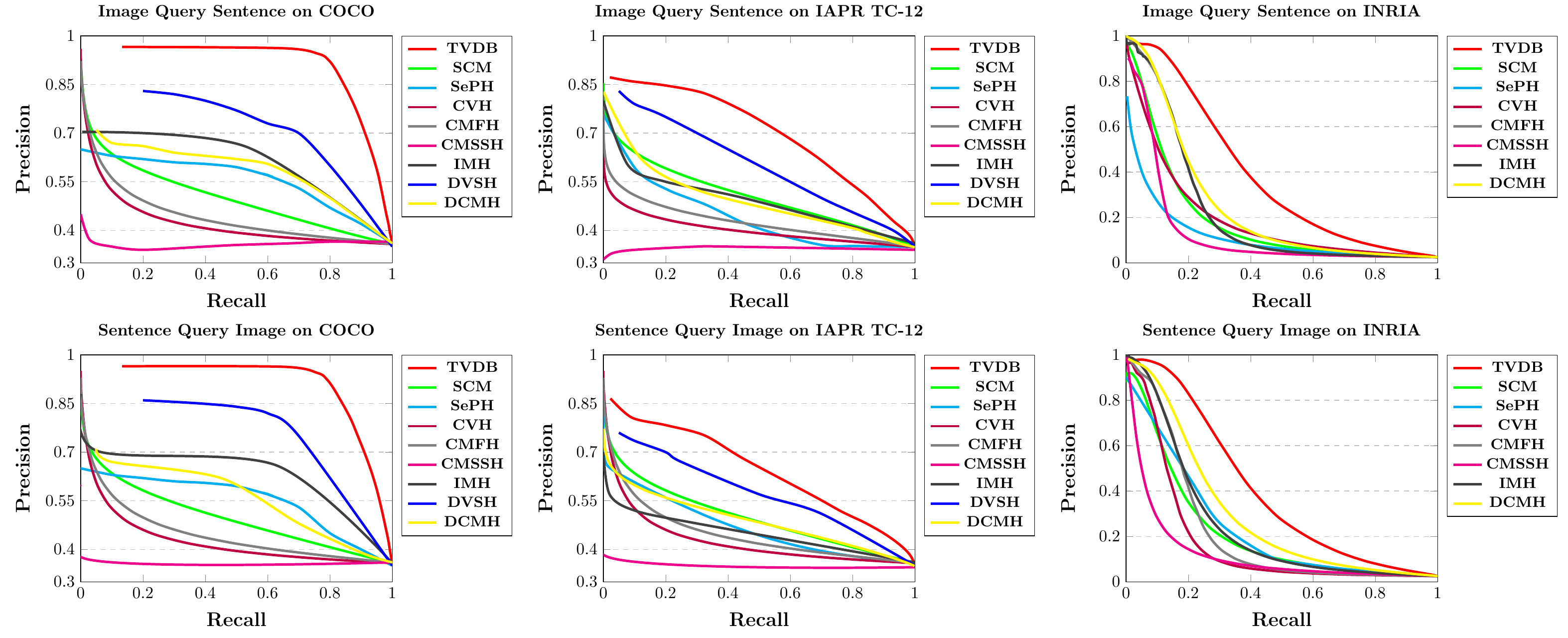}
	\end{center}
	\vspace{-2.5ex}
	\caption{Cross-modal retrieval Precision-Recall curves of TVDB and some existing methods with code length $M=32$.}
	\vspace{-2ex}
	\label{prcurve}
\end{figure*}

\textbf{IAPR TC-12 \cite{iapr}.} It consists of 20,000 images. Each image is provided with 1.7 descriptive sentences on average. In addition, category annotations are given on all images with 275 concepts. Following the setting in \cite{dvsh}, we use 18,000 image-sentence pairs that belong to the most frequent 22 topics as the retrieval gallery, from which we take 2,200 pairs as the query data and 5,000 as the training set.

\textbf{INRIA Web Queries \cite{inria}.} This dataset contains about 70,000 images categorized into 353 conceptual labels. Sentence descriptions are provided to most of the images. We select images belonging to the 100 most frequent concepts, making the gallery 25,015 image-sentence pairs. For the query set, 10 images and sentences are randomly selected from each category and the rest are used as training data.
\subsection{Comparison with existing methods}

The overall retrieval performance of TVDB is analysed and compared using Mean Average Precision (MAP) and Precision-Recall curves. %More comprehensive results are given in the \textbf{supplementary document}.

\vspace{1ex}\textbf{Baselines.} Several baselines of traditional cross-modal hashing methods are adopted for comparison, including CMFH~\cite{cmfh}, CVH~\cite{cvh}, SCM\cite{scm}, IMH~\cite{imh}, QCH~\cite{qch}, CMSSH~\cite{cmssh}, SePH~\cite{seph}, while CVFL~\cite{cvfl}, CCA~\cite{cca} and PLSR~\cite{plsr} are considered as real-valued methods. The deep-learning-based cross-modal hashing methods, \ie, CorrAE~\cite{corrae}, CM-NN~\cite{cmnn}, DNH~\cite{dnhc}, DCMH~\cite{dcmh} and DVSH~\cite{dvsh}, are also included here. To make fair comparisons, we utilize deep features for all traditional baseline methods mentioned above if the codes are available. For image features, we directly use the 4096-D AlexNet~\cite{cnn2} pre-trained representations. 
For text features, a multi-label classification text-CNN, which shares most of its structure with our text encoding network $\mathit{g}\left(\cdot\right)$ excluding the last layer, is pre-trained on each dataset.
Then a 384-D feature is extracted for each sentence from the pooling layers. In implementing DCMH~\cite{dcmh}, we build an identical text coding network to ours, enabling it to handle sentence data. For DNH~\cite{dnhc}, we cite the performances of its variants DNH-C provided in \cite{dvsh} since the same settings are used.

\begin{figure}
	\begin{center}
		\includegraphics[width=0.48\textwidth]{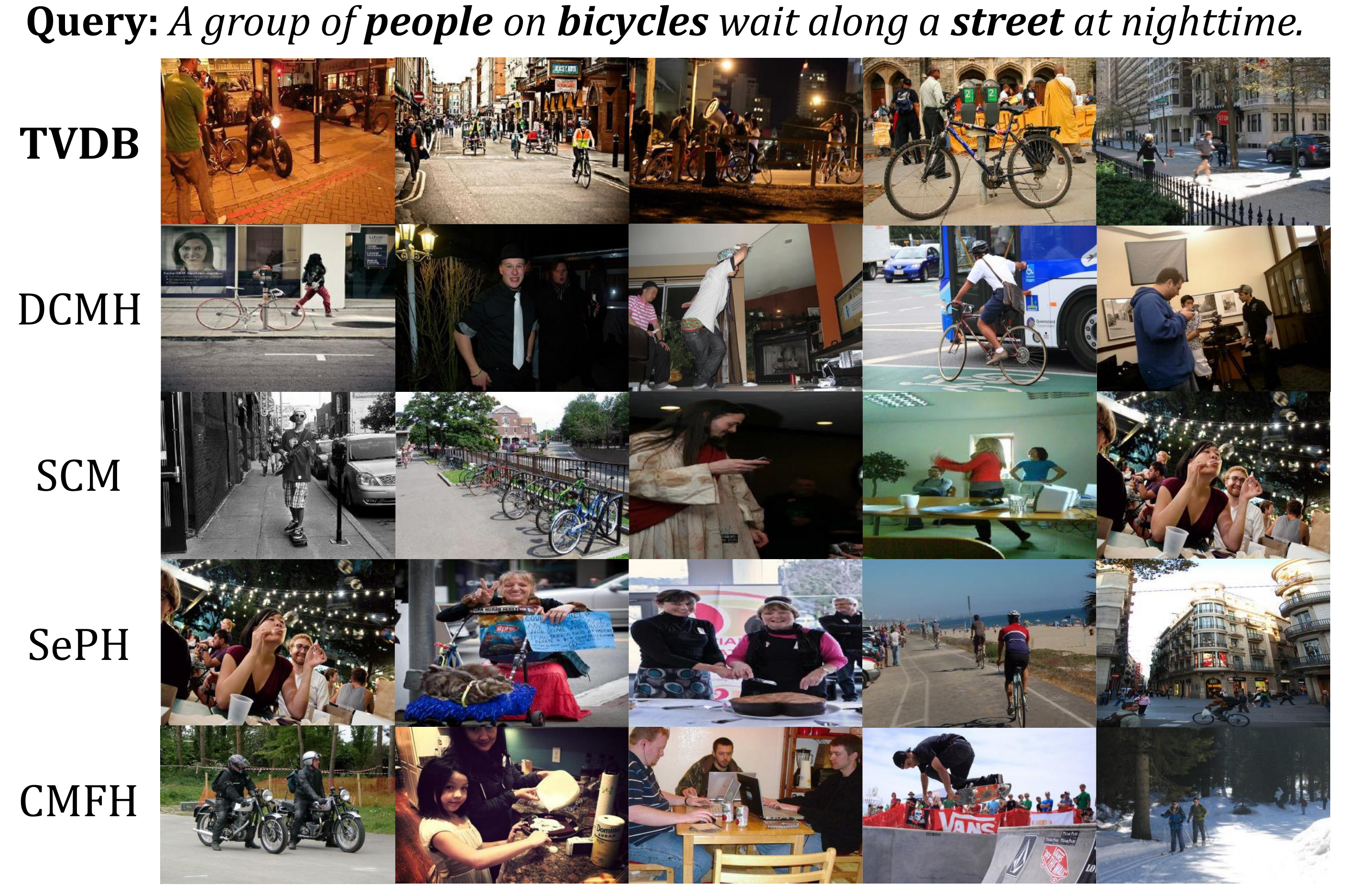}
	\end{center}
	\vspace{-1ex}
	\caption{Intuitive \textit{Sentence Query Image} retrieval top-5 results on Microsoft COCO 128 bits. TVDB carries the best matching candidates, where most of the objects mentioned in the query are included (\eg people, bike, street).}%More examples are provided in \textbf{supplementary document}.}
	\label{hook}
	\vspace{-0.1ex}
\end{figure}

\textbf{Results and Analysis.} The retrieval MAP performances on the three datasets are reported in Table~\ref{cmp1}. In general, the proposed TVDB model outperforms existing methods on the three datasets with large margins. Most traditional cross-modal hashing techniques are not usually designed specifically for images and sentences, which limits their performances on the three datasets. This suggests that image-tag hashing and retrieval are unrepresentative for vision-language tasks. The recent deep hashing model DVSH~\cite{dvsh} hits the closest overall figures to ours as its modal-specific deep networks are able to explore the intrinsic semantics of images and sentences. TVDB provides even superior performance since both regional image information and relations between the words are well encoded, making the output binary codes more discriminative. Some results are left blank because the corresponding baseline codes are not available and the performances are never reported. Note that the overall MAP scores on INRIA Web Queries \cite{inria} are relatively lower than those on the other two. This is probably due to the relatively low image quality. The corresponding precision-recall curves with 32-bit code length are given in Figure~\ref{prcurve}. A \textit{Sentence Query Image} example is given in Figure~\ref{hook} with top-5 closest retrieved candidates on 128-bit COCO. TVDB provides well-matched results with detailed information preserved (\eg people, bike, street), while the compared methods do not.

\subsection{Ablation study of deep encoding network}

We demonstrate the impact of the deep encoding networks for TVDB in this subsection.

\textbf{Baselines.} Four variants of TVDB are built as baselines by modifying the deep encoding networks with some other architectures: \textbf{(a) TVDB-I1} is built by replacing the region-based image encoding network of TVDB by a holistic AlexNet CNN. \textbf{(b) TVDB-I2} mixes the image region features using average pooling instead of rendering them to the LSTM units. \textbf{(c) TVDB-T1} takes text bag-of-words as sentence features with the original text-CNN removed. \textbf{(d) TVDB-T2} is a variant of TVDB where the text-CNN is replaced by a two-layer LSTM structure.%, where images are encoded by a simple AlexNet CNN and the text encoding network is LSTM-based.

\vspace{0.7ex}\textbf{Results and Analysis.} Self-comparison MAP results on cross-modal retrieval are shown in Table~\ref{cmp2}. As we expected, the MAP scores drop dramatically with simple image CNN (TVDB-I1, TVDB-I2) and bag-of-words features (TVDB-T1), but are still acceptable compared to some existing methods. Image binary encoding without regional information is still far from satisfactory. TVDB-T2 with text-LSTM obtains reasonable performance and is in general superior to the state-of-the-art DVSH~\cite{dvsh}, since LSTM is also capable of modeling sentences. However, TVDB-T2 performs poorer than the original TVDB, suggesting that the proposed text-CNN architecture is a suitable choice for the cross-modal hashing task. To this end, it can be seen that our proposed binary encoding network is successfully designed and all components are reasonably implemented.

\subsection{Training efficiency}

The training efficiency and convergency of the proposed \textit{stochastic batch-wise learning routine} are illustrated in this subsection, by comparing it with several baselines.% of TVDB with several other training routines are compared.%, while the networks are not modified for fair comparison with the initial TVDB.

\textbf{Baselines.} \textbf{(a) TVDB-S} varies TVDB by keeping in-batch images and sentences unaltered with each epoch. \textbf{(b) TVDB-N} is a variant of TVDB where $\mathbf{B}\text{ and } \mathbf{H}$ are initialized by $\max_{\mathbf{B}, \mathbf{H}}\operatorname{trace}\left(\mathbf{B}\mathbf{S}\mathbf{H}^\top\right)$ and are not updated during SDG training. Here, $\mathbf{S}$ refers to the similarity matrix on the whole training set as in Eq.~(\ref{eqs}). \textbf{(c) TVDB-E1} is similar to the optimization of DCMH~\cite{dcmh}, performing epoch-wise binary code learning instead of batch-wise, \ie, updating both $\mathbf{B}, \mathbf{H}$ in a similar manner to Eq.~(\ref{updatedb}) and (\ref{updatedh}) on the whole training set after each epoch. \textbf{(d) TVDB-E2} is similar to TVDB-E1, but code learning and updating is performed after every five epochs of training instead of each one.

\begin{figure}
	\begin{center}
		\includegraphics[width=0.47\textwidth]{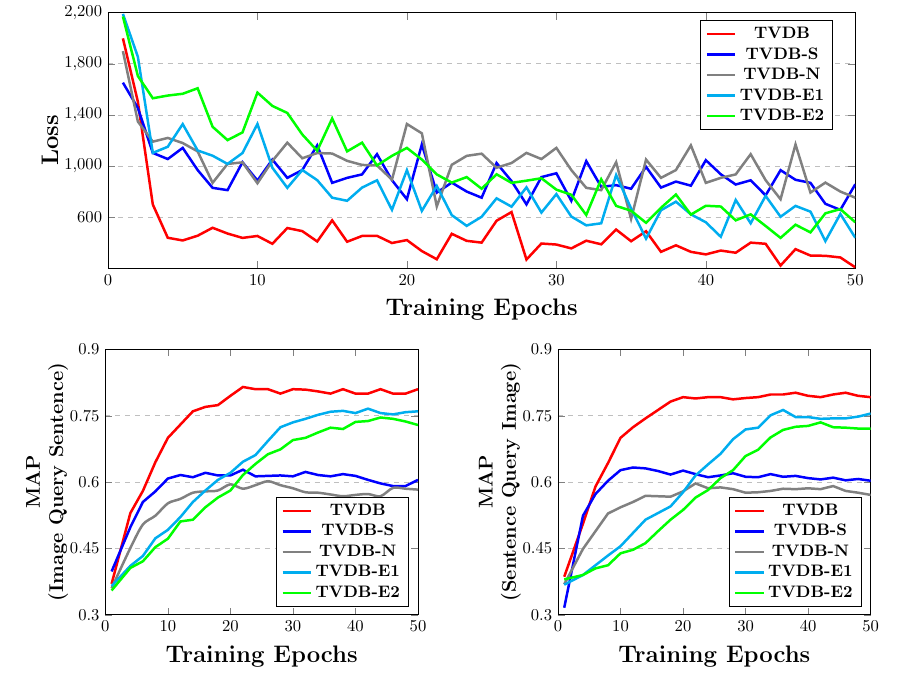}
	\end{center}
	\vspace{-2.5ex}
	\caption{The 128-bit retrieval MAP of \textit{Image Query Sentence} (bottom left) and \textit{Sentence Query Image} (bottom right) \wrt
		training epochs on Microsoft COCO dataset are shown here. The corresponding losses are also given on the top.}
	\label{figeff}
\end{figure}

\textbf{Results and Analysis.} Experiments are conducted on the Microsoft COCO \cite{coco} dataset with code length $M=128$. Cross-modal retrieval MAP scores and corresponding learning losses \wrt training epochs are shown in Figure~\ref{figeff}. It is obvious that TVDB converges quickly to an acceptable MAP score and then gradually hits the best performance at about the 20$^{\text{th}}$ epoch. TVDB is generally superior to the compared baselines both in terms of peak performance and training efficiency. It can be seen that TVDB-N obtains a similar rate of convergency to TVDB for the first five training epochs, but ends up with significantly lower retrieval performance since the network parameters $\varTheta\text{ and }\varPhi$ are disjointly optimized with $\mathbf{B} \text{ and } \mathbf{H}$. TVDB-S follows a close path to TVDB-N with a slightly higher performance. It is clear that our code learning strategy with unaltered in-batch data is not appealing in exploring the generalized optima of $\mathbf{B} \text{ and } \mathbf{H}$ on the whole dataset. TVDB-E1 and TVDB-E2 carry out acceptable retrieval performances but are still outperformed by TVDB. %Although TVDB-E1 and TVDB-E2 also perform alternating optimization, the overall training efficiencies are far from satisfactory. 
This demonstrates that each aspect of TVDB is necessary to obtain optimal performance. %It takes about 35 and 45 epochs to reach the best performance for TVDB-E1 and TVDB-E2 respectively, where TVDB only takes 25 epochs in getting even better performance. %On the other hand, computing $\max_{\mathbf{B}, \mathbf{H}}\text{trace}\left(\mathbf{B}\mathbf{S}\mathbf{H}^\top\right)$ on the whole training set is extremely time consuming, taking 130 minutes for each round for $\mathbf{B} \text{ and } \mathbf{H}$ with an Intel XEON CPU server. However, since Equations~(\ref{updatedb},\ref{updatedh}) are only dealing with in-batch data and can be assigned to GPUs for computation, it requires only about 300ms for each batch of TVDB to work out $\mathbf{B}_b \text{ and } \mathbf{H}_b$, which significantly saves the training time.

%-------------------------------------------------------------------------
\section{Conclusion}
In this paper, we proposed a deep binary encoding method termed as Textual-Visual Deep Binaries (TVDB) which is able to encode information-rich images and descriptive sentences. Two modal-specific binary encoding networks were built using LSTM and text-CNN, leveraging image regional information and semantics between the words to obtain high-quality binary representations. In addition, we proposed a \textit{stochastic batch-wise code learning routine} that performs effective and efficient training. Our experiments justified that both the proposed deep encoding networks and the training routine contribute greatly to the final outstanding cross-modal retrieval performance.
{\small
\bibliographystyle{ieee}
\bibliography{egbib}
}

\end{document}